\documentclass[10pt,twocolumn,letterpaper]{article}
\usepackage{iccv}
\usepackage[utf8]{inputenc} %
\usepackage[T1]{fontenc}    %
\usepackage{url}            %
\usepackage{booktabs}       %
\usepackage{amsfonts}       %
\usepackage{nicefrac}       %
\usepackage{microtype}      %
\usepackage[dvipsnames]{xcolor}         %

\usepackage{colortbl}
\usepackage{graphicx,overpic}
\usepackage[export]{adjustbox}
\usepackage{amsmath, amssymb}
\usepackage{multirow}
\usepackage[super]{nth}
\usepackage{comment}

\usepackage{enumitem}
\usepackage{caption}

\usepackage{tikz}
\usepackage{pgfplots}
\usepackage{pgfplotstable}
\pgfplotsset{compat=1.13}
\usetikzlibrary{pgfplots.groupplots}
\usetikzlibrary{calc,arrows}

\usepackage{tabularx,stackengine,collcell}
\let\endminwd\relax
\newcolumntype{L}[1]{>{\collectcell\xminwd l{#1}}l<{\endminwd\endcollectcell}}
\newcolumntype{C}[1]{>{\collectcell\xminwd c{#1}}c<{\endminwd\endcollectcell}}
\newcolumntype{R}[1]{>{\collectcell\xminwd r{#1}}r<{\endminwd\endcollectcell}}
\def\minwd#1#2#3\endminwd{\stackengine{0pt}{#3}{\rule{#2}{0pt}}{O}{#1}{F}{F}{L}}
\newcommand\xminwd[1]{\minwd#1}

\usepackage{pifont}
\newcommand{\mj}{$\mathcal{J}$}
\newcommand{\mf}{$\mathcal{F}$}
\newcommand{\mjf}{$\mathcal{J}\&\mathcal{F}$}

\newcommand{\mjs}{$\mathcal{J}_s$}
\newcommand{\mfs}{$\mathcal{F}_s$}
\newcommand{\mju}{$\mathcal{J}_u$}
\newcommand{\mfu}{$\mathcal{F}_u$}
\newcommand{\mg}{$\mathcal{G}$}

\definecolor{defaultColor}{RGB}{230, 230, 250}

\usepackage[pagebackref,breaklinks,colorlinks,allcolors=iccvblue]{hyperref}

\usepackage[normalem]{ulem}

\newcommand{\image}{I}
\newcommand{\width}{W}
\newcommand{\wsmall}{w}

\newcommand{\hsmall}{h}

\newcommand{\height}{H}

\newcommand{\channel}{c}

\newcommand\boldblue[1]{\textcolor{blue}{\textbf{#1}}}

\definecolor{iccvblue}{rgb}{0.21,0.49,0.74}
\usepackage[pagebackref,breaklinks,colorlinks,allcolors=iccvblue]{hyperref}

\def\paperID{7729}
\def\confName{ICCV}
\def\confYear{2025}

\title{Structure Matters: Revisiting Boundary Refinement in Video Object Segmentation}

\author{
Guanyi Qin$^{1,}$\thanks{~Equal contribution, $\dagger$~Corresponding author.}~, Ziyue Wang$^{1,}$\footnotemark[1]~, Daiyun Shen$^{1}$, Haofeng Liu$^{1}$, Hantao Zhou$^{2}$\\ Junde Wu$^{3}$, Runze Hu$^{4}$, Yueming Jin$^{1,\dagger}$\\
$^{1}$National University of Singapore; $^{2}$Tsinghua University\\$^{3}$University of Oxford; $^{4}$Beijing Institute of Technology\\
{\tt\small  {\{guanyi.qin, e1374378, e1374467, haofeng.liu\}@u.nus.edu}, hrz@bit.edu.cn, ymjin@nus.edu.sg}
}

\begin{document}
\maketitle
\begin{abstract}
Given an object mask, Semi-supervised Video Object Segmentation (SVOS) technique aims to track and segment the object across video frames, serving as a fundamental task in computer vision. Although recent memory-based methods demonstrate potential, they often struggle with scenes involving occlusion, particularly in handling object interactions and high feature similarity. To address these issues and meet the real-time processing requirements of downstream applications, in this paper, we propose a novel b\textbf{O}undary \textbf{A}mendment video object \textbf{S}egmentation method with \textbf{I}nherent \textbf{S}tructure refinement, hereby named \textbf{OASIS}. Specifically, a lightweight structure refinement module is proposed to enhance segmentation accuracy. With the fusion of rough edge priors captured by the Canny filter and stored object features, the module can generate an object-level structure map and refine the representations by highlighting boundary features. Evidential learning for uncertainty estimation is introduced to further address challenges in occluded regions. The proposed method, OASIS, maintains an efficient design, yet extensive experiments on challenging benchmarks demonstrate its superior performance and competitive inference speed compared to other state-of-the-art methods, i.e., achieving the \mf~values of $91.6$ (vs. $89.7$ on DAVIS-17 validation set) and \mg~values of $86.6$  (vs. $86.2$ on YouTubeVOS 2019 validation set) while maintaining a competitive speed of $48$ FPS on DAVIS. Checkpoints, logs, and code will be available \href{https://github.com/jinlab-imvr/OASIS}{here}.
\end{abstract}     
\section{Introduction}
\label{sec:intro}
Video Object Segmentation (VOS) is an important task in computer vision \cite{early1, davis2016} that aims to separate moving objects from the background. In the general semi-supervised VOS (SVOS) setting, the task involves propagating the object masks specified in the first frame throughout the entire video.
By enabling effective identification and understanding of objects in videos, VOS plays a crucial role in various applications including autonomous driving, robotics, video editing, medical intervention, etc. Albeit performance demands, real-time processing is also essential for these applications, necessitating algorithms that accurately perceive and segment objects while delivering sufficient inference speed.

\begin{figure}[!t]
    \centering
    \includegraphics[width=0.8\linewidth]{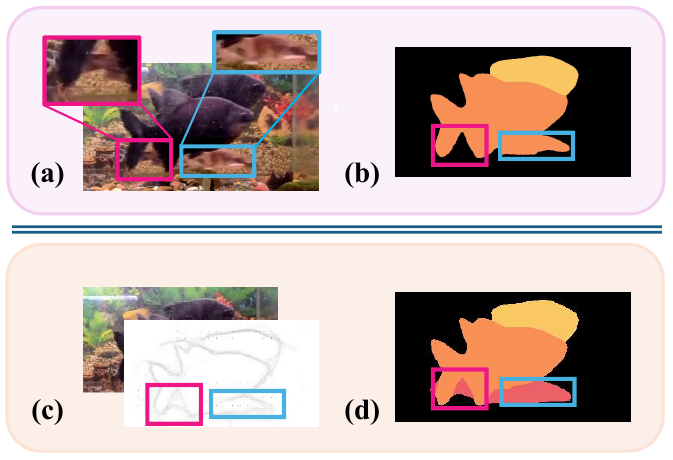}
    \caption{(a) Input frame. (b) Object mask generated by the SOTA method Cutie. (c) Output structure map. (d) Object mask generated by our model. In this case, the fish corresponding to the red mask closely resembles the background and is concealed behind the other two fish, causing previous methods to fail. Guided by our structure map, the model successfully segments the fish accurately.}
    \label{fig:fg-demo}
\end{figure}

Recent advancements \cite{xmem, aot, deaot, deva} in SVOS emphasize the value of memory-based mechanisms. Representations of certain past frames are generated and stored in a memory bank. The current frame is then used as the query to match these stored representations and derive segmentation masks.
Cutie \cite{cutie}, further proposes object queries as a high-level summary of the target object and interacts with the pixel-level features of the query frame for accurate segmentation, achieving the state-of-the-art performance. However, these approaches still struggle when similar objects are close to each other or are occluded, as shown in Figure \ref{fig:fg-demo} (a) and (b).
On one hand, the mask prediction depends predominantly on maintaining consistency across frames for both pixel- and object-level features, while the inherent geometric characteristics of the objects are often overlooked. This limitation poses challenges in accurately discerning object structures, particularly when they resemble the background (red box).
On the other hand, when occlusion occurs, objects become fragmented. This challenge is amplified when objects share similar semantics, leading to high uncertainty in overlapping areas. Without an understanding of object structure, the model struggles to distinguish objects and establish clear boundaries (blue box).

Edge information is a crucial component of an object's structure representation, defining its boundaries and shape by representing the inherent structural information. Precise edge prediction allows models to capture structural details more effectively, thereby enhancing the model's ability to understand occlusions. Therefore, edges have been widely used for segmenting objects for static images across various fields including object detection and semantic segmentation \cite{overview1, overview2, edge1}. However, in VOS tasks, target movements lead to considerable shifts in object edges over time, making it challenging to predict accurate edges from scratch for each frame. This often requires large edge detection networks, which significantly reduce inference efficiency. Therefore, edges are only used to analyze motion magnitude or shadow area detection primarily, which merely requires coarse edges \cite{video4, video5, evs3}. 

To tackle the aforementioned problems, we propose an b\textbf{O}undary \textbf{A}mendment video object \textbf{S}egmentation with \textbf{I}nherent \textbf{S}tructure refinement, hereby named \textbf{OASIS}, which effectively refines object structures to handle geometric and layer relationships among objects better, while highlighting them out of backgrounds.
This is achieved through a simple but effective structure refinement module. 
Specifically, we first extract the rough edge information by utilizing a fast and robust edge detection algorithm, Canny Edge Detection \cite{canny1}, which can identify global edges in individual frames and provide strong priors. With these priors, we only need a lightweight structure decoder to obtain the edges of the target objects instead of predicting from scratch without significant sacrifice in inference speed. This map further amplifies the original features by highlighting objects' boundaries, producing structure-enhanced features for final segmentation. This process enables the model to differentiate among objects and from the background effectively.
Additionally, an object feature fusion mechanism is introduced to integrate memory features into the structure map to better model relationships between objects. 
Furthermore, to address the high uncertainty in dealing with overlapped areas, we also make the first attempt to deliver a new learning pattern that incorporates Evidential Learning (EDL) for uncertainty quantification and introduces an additional loss to minimize uncertainty and improve segmentation accuracy in occlusion regions.

Extensive experiments demonstrate the superiority of our approach across four benchmarks under two training settings, particularly in scenes where target objects interact and exhibit complex relationships. 
Notably, OASIS achieves promising results, consistently outperforming current state-of-the-art methods across different datasets, while delivering a real-time inference speed at 48 fps.
In summary, our contributions are the following:
\begin{itemize} 
\item We propose OASIS, a novel framework designed to manage relationships between the objects and the background by accurately modeling object structures. 
\item We introduce a structure refinement module and an object sensory mechanism that utilizes edge information to achieve inherent structure refinement. 
\item To the best of our knowledge, we make the very first attempt to incorporate evidential learning in the SVOS task to improve the model performance on high-uncertainty regions, particularly in areas with significant similarity and occlusion. 
\item We verify OASIS on multiple SVOS benchmarks involving a wide range of video contents, annotation conditions, and training settings. OASIS outperforms other competitors across all these datasets and settings.
\end{itemize}
\section{Related Work}
\label{sec:rel_work}

\subsection{Video Object Segmentation}
\label{subsec:re_vos}
Early approaches of VOS \cite{davis2016,davis2017,early1,early2,monet,premvos,maskrnn,early3,early4} implement online learning that finetunes the model at test-time, resulting in low efficiency. 
Propagation-based methods \cite{segflow, prop1, prop2, early3, prop3, prop4, prop5, sstvos, prop6, prop7} iteratively propagate masks between adjacent frames by leveraging temporal correlations; however, they are inevitably constrained by drift and occlusion. Matching-based models \cite{match1, match2, match3, feelvos, stm, match5, stcn, match7, match8, match9} compute correspondences between the current and reference frames to capture long-range temporal context. Some memory-based methods \cite{xmem, match5, stcn, match8, memory1, stm, memory2, memory3, prop4, memory4, memory5, resurgsam2,TMIIVT} further enhance this by using memory banks to store features from past frames for pixel-level matching. The lately research SAM2 \cite{sam2} also benefitted from memory mechanisms.
Recently, Cutie \cite{cutie} proposed an object-reading mechanism to enhance object-level representations. While it effectively perceives individual objects, modeling the relationships between multiple objects remains an open challenge.

\begin{figure*}[!ht]
    \centering
    \includegraphics[width=0.95\linewidth]{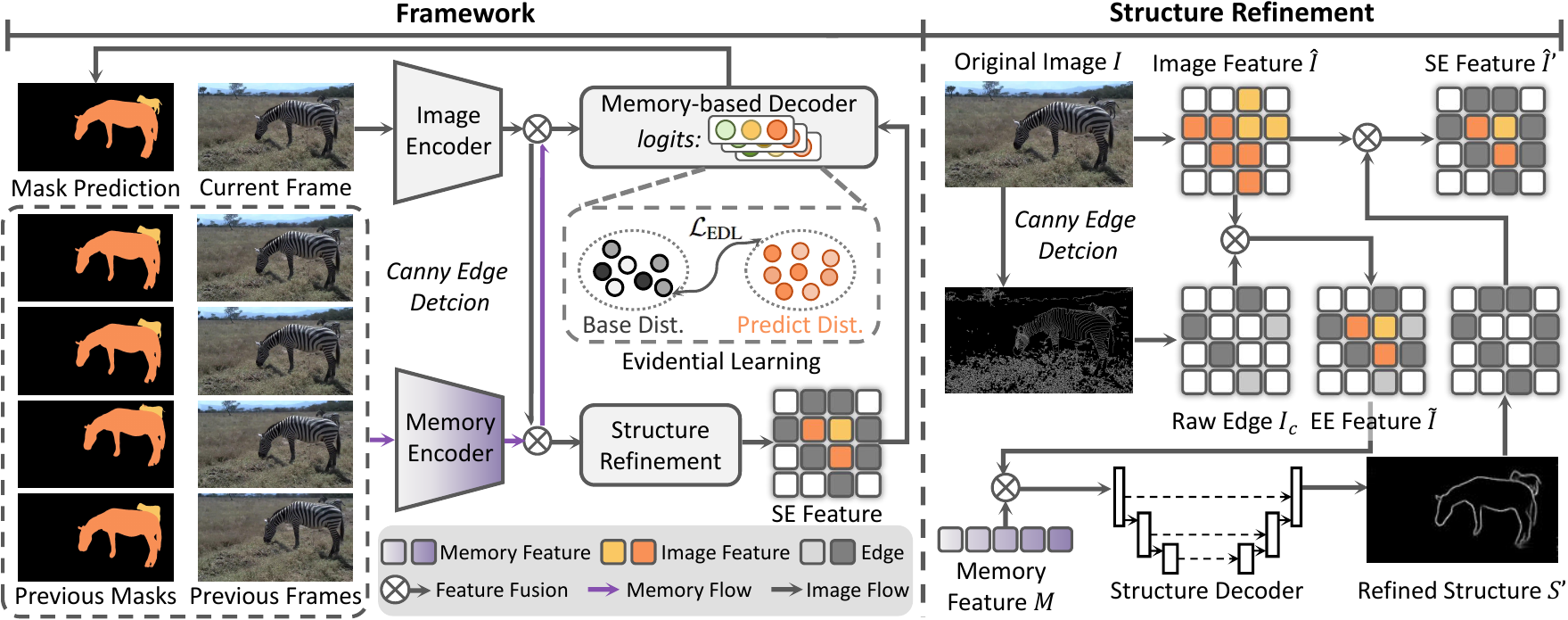}
    \caption{Overview of our proposed method: the left side shows the overall framework, while the right side provides a detailed illustration of the structure refinement process. The light grey box on the left indicates evidential learning, and "Dist." is an abbreviation for "Distribution." }
    \label{fig:net-overview}
\end{figure*}

\subsection{Edge Detection}
\label{subsec:re_ed}
Edge information has been widely utilized to differentiate occluded objects \cite{overview1, overview2, edge1, ziyue1, ziyue2, ziyue3}. In salient object detection \cite{det1, det2, det3, det4, det5, det6}, edge features assist in identifying the most visually prominent objects within the image. Similarly, in segmentation tasks \cite{seg1, seg2, seg3, video3, video4, video5}, edge detection helps to distinguish different regions and instances. Accurate localization of moving objects in videos remains a significant challenge, with edge information playing a crucial role in mitigating this difficulty. \cite{real-time1, video1, video2, video3, video4, video5}. However, these methods require complex network designs like extra convolution layers, greatly affecting inference efficiency. Canny edge detection \cite{canny1,canny2,canny3,canny4} is a popular algorithm for identifying sharp intensity changes in images, which can help identify the edges of all regions, making it a valuable prior for enhancing edge prediction. \cite{cannyapp1,cannyapp2,cannyapp3,cannyapp4}.

\subsection{Video Segmentation with Edge Information}
\label{subsec:re_evos}
Edge information can effectively capture object movements, offering significant potential for advancing video object segmentation. Studies such as \cite{video4, video5} demonstrated that edge information can aid in analyzing motion magnitude in videos. \cite{video3} further enhances this by combining flow edges with optical flow to track moving objects. Moreover, edge information has been utilized for shadow segmentation in videos \cite{evs1}, while \cite{evs3} leverages edge information to derive discriminative representations, facilitating the differentiation of various semantic regions. However, these approaches either operate solely at the image level without distinguishing objects or are unable to generate fine-grained segmentation results for object tracking, rendering them unsuitable for semi-supervised video object segmentation.

\subsection{Uncertainty Estimation}
\label{subsec:edl}
Uncertainty estimation involves predicting the level of confidence or uncertainty in a network’s output, encompassing aleatoric uncertainty (arising from data noise) and epistemic uncertainty (stemming from model limitations) \cite{edl2}. Early approaches employed methods like particle filtering and CRFs \cite{edl8, edl9}. With the rise of deep learning, Bayesian neural networks \cite{edl5, edl11, edl12} were introduced, along with other techniques such as ensemble methods \cite{edl6}. More recently, evidential learning \cite{edl, edl2, edl3, edl4} has shown promising results, which has the potential to improve the high uncertainty of occlusion areas in VOS tasks.

\section{Methodology}

\subsection{Inherent Structure-guided SVOS}
\noindent \textbf{Problem Formulation} Given an object mask for reference, existing memory-based SVOS methods typically learn to segment objects sequentially across frames while concurrently updating the model memory for future regards. This procedure can be expressed as:
\begin{equation}\label{eq:m1}
\begin{gathered}
    q_t = h(\varphi(i_{t}),y_{t}), \\
    y_t = \varrho(i_{t-1}, \dots,i_{t-n}|q_{t-1}, \dots,q_{t-n}),
\end{gathered}
\end{equation}
where $ i_{t} $ denotes the input frame and $ y_{t} $ is the constructed memory from previous $n$ frames and predictions, acting as reference at time $ t $. Functions $ \varphi $ and $ \varrho $ represent the feature extractor and memory constructor, respectively. The notation $ h $ refers to the object segmentation network, which utilizes frame features $ \varphi(i_{t}) $ and memory $ y $ to produce the segmentation output $ q $ at time $ t $. This pipeline interprets the SVOS task from the perspective of the frame content.

However, we observe that Eq.~\ref{eq:m1} lacks adequate modeling of structural information and hierarchy relationships. To enhance the capacity of SVOS and better address issues of occlusion and similarity-induced confusion, we introduce a novel structure refinement module to the existing framework. This module provides the model with inherent structural information and hierarchical relationships in each input frame. Fig. \ref{fig:net-overview} shows the overall architecture of our proposed model. 

Specifically, given an input frame $ i_{t} $, we first derive its edge representation using the Canny filter $c$, a robust algorithm known for accurately detecting edges despite the noise. The frame features $ \varphi(i_{t}) $ then interact with the edge representation to generate an inherent structure map $ s $ under the guidance of the memory features, capturing key structural characteristics of target objects. This inherent structure map is subsequently integrated back into the frame features. Finally, we align memory features and structured enhanced features to predict the target object mask. Accordingly, our proposed solution can be formulated as:
\begin{equation} \label{eq:m2}
\begin{gathered}
    q_t = h(\varphi(i_{t}), y_{t}, s_{t}), \\
    s_{t} = \vartheta(\varphi(i_{t}), c(i_{t}), y_{t}),
\end{gathered}
\end{equation}
where $\vartheta$ denotes the fusion function for processing features and inherent structure map. By explicitly modeling structural information, our approach allows the model to better distinguish objects, even in challenging scenarios involving occlusion and similar semantics.

\subsection{Inherent Structure Map Generation}

\noindent \textbf{Edge Highlighting} Given an input RGB image $ \image \in \mathbb{R}^{3 \times \height \times \width} $, we first derive its edge representation $ \image_{c} \in \mathbb{R}^{1 \times \height \times \width} $ using the Canny filter. Next, $ \image $ is fed into an image encoder for multi-scale feature extraction. The output of the image encoder can be formulated as $ \hat{\image}_{i} \in \mathbb{R}^{\channel_{i} \times \hsmall_{i} \times \wsmall_{i}} $, for $ i = 1, \dots, N $, where $ N $ denotes the number of feature scales, $\hsmall_{i} = \frac{\height}{2^{i+1}}$ and $\wsmall_{i} = \frac{\width}{2^{i+1}}$.

Since the image encoder is pre-trained on ImageNet~\cite{imagenet}, its output feature map tends to emphasize abstract, content-focused information. In the case of SVOS, this nature may impede the model to accurately segment objects, potentially leading to confusion when occlusions occur among fragmented objects. To enhance the sensitivity of model to edges and structural details, we incorporate edge representation from $ \image_{c} $ into $ \hat{\image}_{{i}} $ through an element-wise multiplication, highlighting edge out of the feature map. This procedure can be formalized as:
\begin{equation}
\label{eq:roughfusion}
\tilde{\image}_{i} = \hat{\image}_{i} + (\hat{\image}_{i} \odot \epsilon \image_{c}^{\downarrow \hsmall_{i}, \wsmall_{i}}), \text{for} \ i = 1, \dots, N,
\end{equation}
where $ \tilde{\image}_{i} $ denotes the global-edge-enhanced feature map, $\odot$ represents the Hadamard product, $\epsilon$ is an importance factor that modulates the influence of the edge map detected by the Canny filter, and $ \image_{c}^{\downarrow \hsmall_{i}, \wsmall_{i}} $ indicates that $ \image_{c} $ has been interpolated to match the spatial dimensions of $ \hat{\image}_{i} $.

\noindent \textbf{Memory Feature Extraction} 
Given the $t$-th frame of the video, we choose the previous $n$ frames and their corresponding predicted masks and feed them into the memory encoder to derive the global memory feature $N$. Memory encoder is a common-used module in SVOS, where we adopt the design from Cutie \cite{cutie}. 
We then project global features onto the object-based masks to derive object memory features $M \in \mathbb{R}^{C\times H\times W}$ (with $H=W=30$ here), which provides temporal information about the target object for the next structure refinement module.

\noindent \textbf{Supervision Preparation} 
Although the global-edge-enhanced feature map $\tilde{\image}_{i}$ emphasizes edges, it still contains boundary details of non-target objects, which is insufficient for guiding the model toward precise segmentation. To address this limitation, we design a structure decoder to predict structure map $S$, containing only edges and structure details of targets, from the global-edge-enhanced feature $ \tilde{\image}_{i} $. For supervison, the ground-truth structure map is derived from ground-truth mask. Given a colored mask $G \in \mathbb{R}^{3 \times H \times W}$ corresponding to the image $\image$, we first convert it into a grayscale image $G_{gr} \in \mathbb{R}^{1 \times H \times W}$. Subsequently, a predefined filter is applied to transform ${G}_{gr}$ into the final object-related inherent structure map $S$. This generated object structure map effectively highlights occlusion relationships among target objects and is further used as ground truth in structural extraction network training.

\begin{figure}
    \centering
    \includegraphics[width=0.92\linewidth]{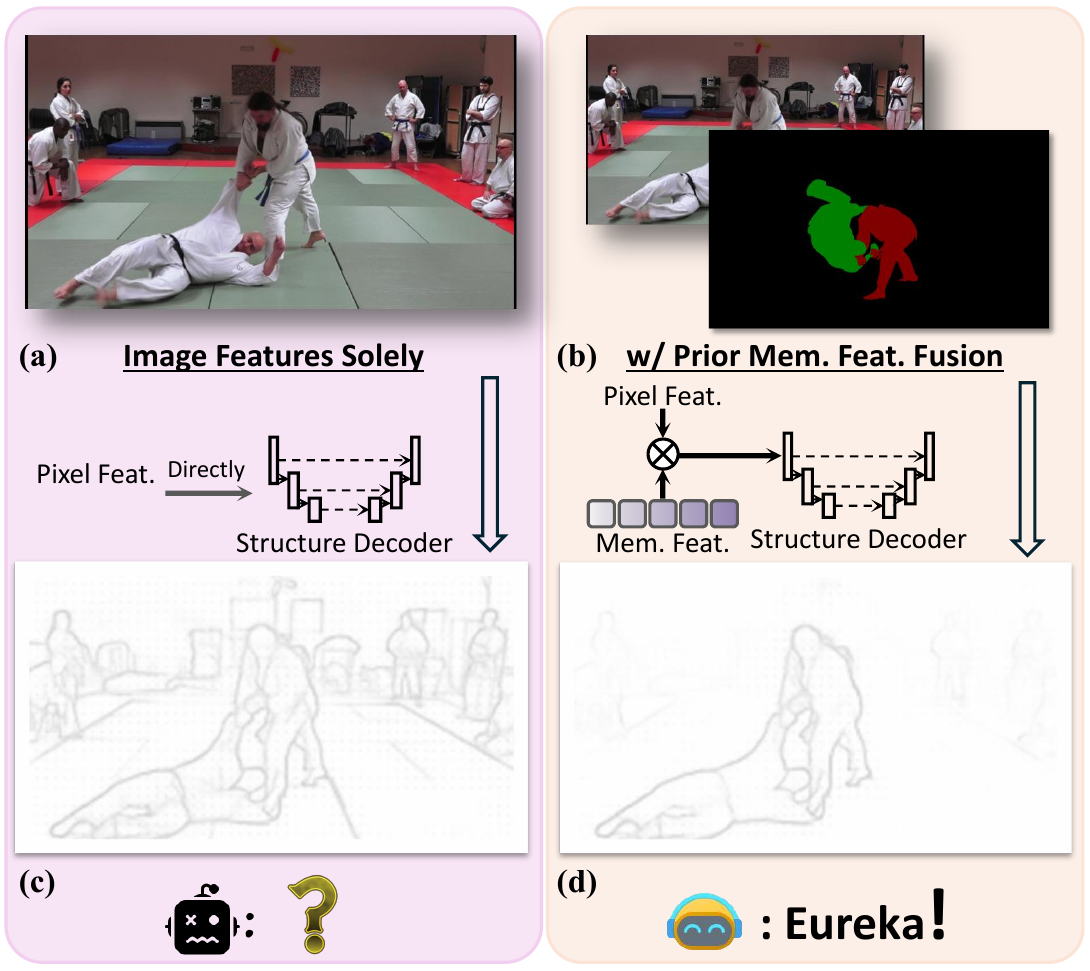}
    \caption{Output structure maps from different objects. (a) Input frame. (b) Input frame and predicted masks of the previous frame. (c) Output structure map without object fusion. (d) Output structure map with object fusion. "Feat." and "Mem." are abbreviations of "Feature" and "Memory" respectively.}
    \label{fig:sensory}
\end{figure}

\noindent \textbf{Object Structure Map Prediction}
After obtaining the global-edge-enhanced feature map $ \tilde{\image}_{i} $ and the object memory feature $M$ from the previous framework, a lightweight structure decoder with connecting paths is applied to predict structure map $S$. 
Importantly, $ \tilde{\image}_{i} $ alone provides limited target-specific cues, and the network could only rely on global image features, which may compromise the accuracy of the target objects' structure map prediction.
Therefore, the fusion of the object memory feature $M$ to incorporate object-specific information is necessary.
This process can be formulated as:
\begin{equation} \label{eq:m3m}
\begin{gathered}
S^{\prime}=\text{Structure\space Decoder}(\tilde{\image}_{i}, M),\ \text{for} \ S^{\prime} \rightarrow S,
\end{gathered}
\end{equation}
where $M$ is fused with the coarsest scale $\tilde{I}_3$ via element-wise addition. The decoder progressively upsamples the fused features with transposed convolutions, incorporating higher-resolution features via channel concatenation. To reduce overhead, binary cross-entropy loss is applied only at the final output.
Visual analysis in Fig.~\ref{fig:sensory} reveals that incorporating object memory features is crucial for accurately capturing the target objects' structure. Otherwise
, the network lacks focus, leading to noisy predictions.

\subsection{Refinement and Decoding}

After obtaining the final object-level structure map, we apply the Hadamard product with the extracted image features to emphasize object boundary features, thus to enhance feature with structure details. This process can be expressed as follows:
\begin{equation}
\label{eq:refine}
\hat{\image}_{i}^{\prime} = \hat{\image}_{i} + (\hat{\image}_{i} \odot \beta S^{\prime~\downarrow \hsmall_{i}, \wsmall_{i}}), \text{for} \ i = 1, \dots, N,
\end{equation}
where $\hat{\image}_{i}^{\prime}$ denotes the features ultimately used to segment the object mask and $\beta$ is an importance factor that modulates the influence of the structure map. Note that the structure map used in this process is in the form of logits rather than a binary mask, which helps to emphasize certain regions when the structure decoder has higher confidence in these areas being boundaries.

With the structure-enhanced feature (SE feature) of the current frame, the global memory features $N$, and the object memory features $M$, we employ a decoder to align them and output the final object-based mask. The memory-based decoder design is inspired by Cutie \cite{cutie}, a pioneering approach to enhance VOS performance by object memory modeling. Concretely, the decoder integrates top-down object features with bottom-up global features so that a dynamic interaction between high-level and low-level representations is established. The SE feature, in this way, serves as an additional pathway for connecting high-level and low-level representations, as the structure encodes object information and latently emphasizes the original feature map accordingly. 

\subsection{Enhancement via Uncertainty Quantification}
Due to the LogSoftmax used in the cross-entropy loss function, the predicted class probabilities remain point-based estimations that collapse multiple types of uncertainty into overconfident predictions~\cite{edl, edl_med}. To address this, we further borrow the ideas of Evidential Deep Learning (EDL) to add an uncertainty-related loss term into our training pipeline, as being supplementary to existing loss combination, cross-entropy loss $ \mathcal{L}_{\text{CE}} $ and dice loss $ \mathcal{L}_{\text{Dice}} $, to help improve performance by quantifying and minimizing the uncertainty of predictions, which is especially crucial in challenging scenarios involving occlusions and features with high similarity.

The EDL based on the theory of subjective logic is used to calculate uncertainty with the Dirichlet distribution which acts as an updating measure of the model's intrinsic prediction bias and provides the reference for the subsequent $\text{KL}_{\text{div}}$ regularization. 
Before training, we initialize a bare Dirichlet $ D(\mathbf{p}|\boldsymbol{\alpha})$ over the class probability vector $\mathbf{p}\!=\![p_1,\dots,p_k]$ and set all $\boldsymbol{\alpha}\!=\!1$, which by Gibbs' inequality means the distribution of maximum uncertainty before seeing any data. During training, we update Dirichlet parameters with the model's output prob-logits to represent updated intrinsic bias.
For each pixel at frame $t$, we define $q_t^{k}$ as the Boolean ground‐truth label of a certain pixel of $k$-th class at time $t$, and \textit{raw output logit} $\dot q_t$ is $k$-class prob-logit vector of the pixel output by the model.
The evidential loss $ \mathcal{L}_{\text{EDL}} $ is calculated as:
\begin{equation}\label{eq:edl_real}
\begin{gathered}
\mathcal{L}_{\text{EDL}} = \sum_{k=1}^{K} q_t^{k} \cdot \left( \psi(\sum \boldsymbol{\alpha}) - \psi(\boldsymbol{\alpha}) \right)  + \text{KL}_{\text{div}}, \\
\text{for}~\boldsymbol{\alpha} = 1 + \eta(\dot{q_t}),
\end{gathered}
\end{equation}
where $ \boldsymbol{\alpha} $ means the Dirichlet parameter, $\eta$ is the confidence function and $ \psi $ denotes the digamma function, and $\text{KL}_{\text{div}}$ is derived corresponding to \cite{edl}. Subsquently, the total loss for mask predictions $ \mathcal{L}_{\text{mask}} $ during training can be defined as:
\begin{equation}\label{eq:loss_com}
\mathcal{L}_{\text{mask}} = \mathcal{L}_{\text{CE}} + \mathcal{L}_{\text{Dice}} + \lambda \mathcal{L}_{\text{EDL}},
\end{equation}
where $ \lambda $ is the hyperparameter to balance the contribution of the uncertainty loss term. With this combination of loss functions, our model benefits from the discriminative power of $ \mathcal{L}_{\text{CE}} $ and $ \mathcal{L}_{\text{Dice}} $ while leveraging $ \mathcal{L}_{\text{EDL}} $ to address uncertain predictions, thus yielding improved segmentation performance.
\section{Experiments}

\subsection{Implementation Details}

\noindent \textbf{Network Configuration} We employ two ResNet-18 models pretrained on ImageNet to extract image features and encode the mask, respectively. The dimension of the object feature is set to $256$. For the structure decoder, we progressively upsample with each step employing a transposed convolution kernel of size $2$ and a stride of $2$ to achieve $2\times$ upsampling. LeakyReLU is used as the activation function and residual connections are applied within this module. For the prediction layer, a $1\times1$ convolution is applied to output the binary segmentation. Due to space limitations, further details are in the supplementary materials.

\begin{table*}[!ht]
    \centering
    \resizebox{\linewidth}{!}{
        \begin{tabular}{lc||ccc||cccccc||cccccc}
        \toprule[1.5pt] &  & \multicolumn{3}{c||}{MOSE} & \multicolumn{3}{c}{DAV-17 Val} & \multicolumn{3}{c||}{ DAV-17 Test } & \multicolumn{5}{c}{ YouTubeVOS-2019 Val} \\
        \cmidrule{3-16}
        Method & Publication & \mjf & \mj & \mf & \mjf & \mj & \mf & \mjf & \mj & \mf & \mg & \mjs & \mfs & \mju & \mfu \\
        \midrule[1pt]
        STCN \cite{stcn} & NeurIPS'21 & 52.5 & 48.5 & 56.6 & 85.4 & 82.2 & 88.6 & 76.1 & 72.7 & 79.6 & 82.7 & 81.1 & 85.4 & 78.2 & 85.9 \\
        AOT-R50 \cite{aot} & NeurIPS'21 & 58.4 & 54.3 & 62.6 & 84.9 & 82.3 & 87.5 & 79.6 & 75.9 & 83.3 & 85.3 & 83.9 & 88.8 & 79.9 & 88.5 \\
        RDE \cite{rde}  & CVPR'22 & 46.8 & 42.4 & 51.3 & 84.2 & 80.8 & 87.5 & 77.4 & 73.6 & 81.2 & 81.9 & 81.1 & 85.5 & 76.2 & 84.8 \\
        XMem \cite{xmem} & ECCV'22 & 56.3 & 52.1 & 60.6 & 86.2 & 82.9 & 89.5 & 81.0 & 77.4 & 84.5 & 85.5 & 84.3 & 88.6 & 80.3 & 88.6 \\
        LLB \cite{llb} & AAAI'23 & - & - & - & 84.6 & 81.5 & 87.7 & - & - & - & 83.6 & 81.7 & 86.5 & 79.2 & 87.0 \\
        ISVOS \cite{isvos} & CVPR'23 & - & - & - & \boldblue{87.1} & 83.7 & \boldblue{90.5} & 82.8 & 79.3 & 86.2 & 86.1 & 85.2 & 89.7 & 80.7 & 88.9  \\
        DeAOT-R50 \cite{deaot} & NeurIPS'22 & 59.0 & 54.6 & 63.4 & 85.2 & 82.2 & 88.2 & 80.7 & 76.9 & 84.5 & 85.6 & 84.2 & 89.2 & 80.2 & 88.8 \\
        DEVA \cite{deva} & ICCV'23 & 60.0 & 55.8 & 64.3 & 86.8 & 83.6 & 90.0 & 82.3 & 78.7 & 85.9 & 85.5 & 85.0 & 89.4 & 79.7 & 88.0 \\
        Cutie \cite{cutie} & CVPR'24 & \boldblue{61.2} & \boldblue{57.2} & \boldblue{65.3} & 86.8 & \boldblue{83.8} & 89.7 & \boldblue{82.8} & \boldblue{79.2} & \boldblue{86.5} & \boldblue{86.2} & \boldblue{85.3} & \boldblue{89.6} & \boldblue{80.9} & \boldblue{89.0} \\
        \midrule
        \textbf{OASIS} & - & \textbf{62.1} & \textbf{57.8} & \textbf{66.3} & \textbf{88.3} & \textbf{85.0} & \textbf{91.6} & \textbf{83.2} & \textbf{79.3} & \textbf{87.0} & \textbf{86.6} & \textbf{85.7} & \textbf{90.1} & \textbf{81.2} & \textbf{89.4} \\
        \bottomrule[1.5pt]
        \end{tabular}}
    \caption{Overall performance comparison on VOS benchmarks. Numbers are taken from \cite{llb, isvos, cutie}. For the current state-of-the-art method, Cutie~\cite{cutie}, we conduct training from scratch using its publicly available code and re-evaluate its performance on the benchmarks. Boldface and bold blue text indicate the best and second-best performance in each column, respectively.
    }
    \label{tab:main-results}
\end{table*}

\noindent \textbf{Training Pipeline} Similar to previous work, we divide the training into two stages. First, OASIS is pre-trained on static image datasets~\cite{sta1, sta2, sta3, sta4, sta5} with a 3-frame synthesis following \cite{cutie}. For these images, we apply mirroring, random affine transformations cut-and-paste as data augmentation and combine different images to create pseudo-video sequences. Default cropping size and total training iteration are set to $384\times384$ and 80K, respectively. The batch size is set to $16$, and point supervision is utilized with a total of $8192$ sampled points. The learning rate is fixed at $1.0 \times 10^{-4}$, and the max norm of gradient clipping is set to $3$. Notably, due to noise introduced by pseudo-video sequences, the structure decoder and EDL loss component are not included at this stage for the purpose of robustness. 

For the main training stage, the batch size is set to $16$, and is conducted for a total of 125K iterations. The base learning rate is initialized at $1.0 \times 10^{-4}$, with the backbone learning rate scaled by a factor of $0.1$. A step scheduler is employed to control the learning rate, reducing it by a factor of $0.1$ at 100K and 115K iterations. The sequence length is set to 8, with each frame cropped to a size of $480 \times 480$ pixels. The EDL loss coefficient $ \lambda $ is set to $0.01$ with an additional annealing factor to gradually modify the impact of the loss. For the Canny filter, the lower and upper thresholds are set to $50$ and $200$, respectively, and an $L_2$ regularization term is applied to enhance accuracy. Parameters $\epsilon$ and $\beta$ are set to $0.5$ and $1.0$ for rough edge and structure map feature fusion, respectively. The number of sampled points for the loss calculation is increased to $12544$. We implement the model using PyTorch and train it on four A5000 GPUs.

\subsection{Benchmark Datasets and Evaluation Metrics}
In our experiments, several commonly used SVOS datasets are used to verify OASIS, including DAVIS-2017~\cite{davis2016}, YouTubeVOS 2019~\cite{xu2018youtube}, and MOSE~\cite{MOSE}. Specifically, DAVIS-2017 consists of 60 densely annotated videos in the training set, with an additional 30 videos in both the validation and test-dev sets for evaluation. YouTubeVOS-2019 contains 3978 videos sourced from video classification datasets, divided into 3471 for training and 507 for validation. MOSE, known for its crowded environments, increased occlusions, and higher rates of object disappearance, comprises 1507 videos for training, with 311 videos in the validation set. Following previous studies, we conduct experiments using various training dataset settings and reported the results.

In our experiments, three commonly used criteria, Jaccard index \mj, contour accuracy \mf \, and their average \mjf \ are employed to quantify the performance of our proposed method. Among them, the Jaccard index assesses the area-based accuracy of predictions, whereas the contour accuracy measures the localization accuracy of object boundaries. Specifically, the YouTubeVOS evaluation provides detailed results for \mj \ and \mf \ on both seen and unseen objects, and \mg \ is the average of all objects, enabling a more comprehensive assessment of the generalization ability of a model. \mj, \mf, \mjf \ and \mg \ range from 0 to 1 and the reported numbers in our paper are scaled for better readability and formatting, following previous SVOS works \cite{xmem, deva, cutie}. A superior performance results in values closer to 100. Frames per second (FPS) is also employed in ablation study to measure the efficiency of a model.

\subsection{Overall Performance Comparison}
Our experiments compared OASIS with nine representative or state-of-the-art memory-based SVOS methods. Among them, STCN \cite{stcn} and LLB \cite{llb} focus on achieving more accurate matching between query frames and memory banks. RDE \cite{rde}, XMem \cite{xmem}, and DEVA \cite{deva} aims to reduce memory usage and improve efficiency, while AOT \cite{aot}, DEAOT \cite{deaot}, ISVOS \cite{isvos}, and Cutie \cite{cutie} emphasize enhanced object modeling. The analysis of experiments is as follows. Note that, the inference results of MOSE, YouTubeVOS, and DAVIS test-dev benchmarks are then packaged and uploaded to the evaluation server to obtain final results.

\begin{figure*}[!ht]
    \centering
    \hspace{-5mm}\includegraphics[width=0.92\linewidth]{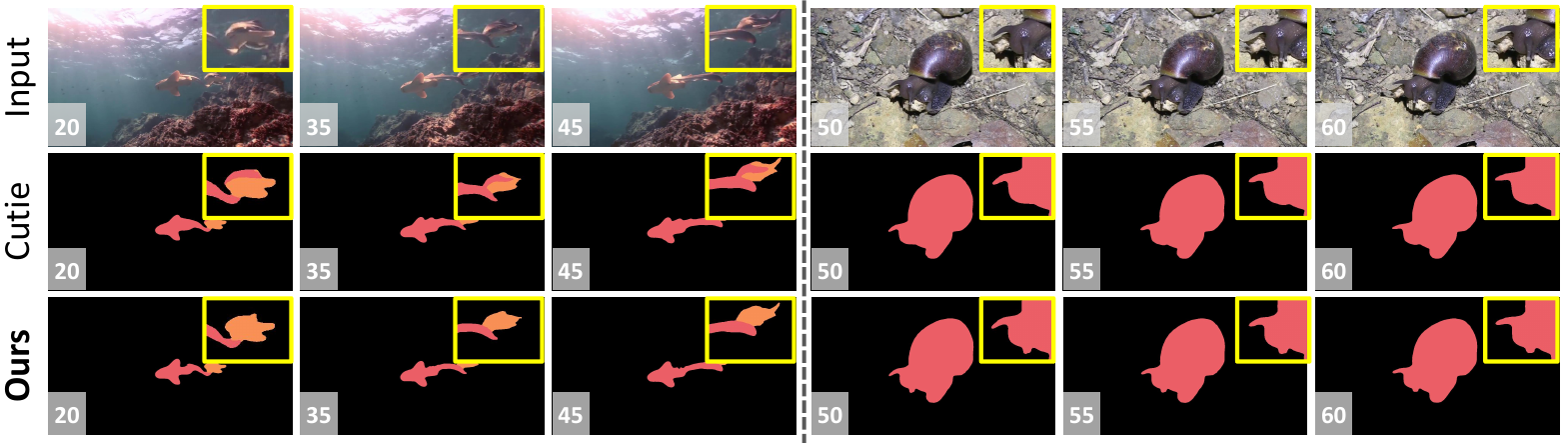}
    \caption{Comparison of segmentation results on two video clips, with details zoomed in on the upper right corner. Notably, our model demonstrates superior resolving capability in occlusion regions and when the target exhibits high similarity to the background. }
    \label{fig:listin}
\end{figure*}

\subsubsection{Quantitive Experiments}
Following the setup in previous works \cite{deva, cutie}, we trained OASIS using a combination of the DAVIS-2017 and YouTube 2019 training sets and evaluated its performance across four validation sets.
It can be seen from Tab.~\ref{tab:main-results} that among all of these SVOS models, our OASIS consistently achieves the best segmentation performance across all benchmarks. Since these four standard benchmarks cover diverse video content and annotation conditions,  \emph{i.e.}, different levels of granularity, achieving top performance on all of them is particularly challenging. These results, therefore, underscore the effectiveness and stability of OASIS. 
Specifically, for datasets with finer annotation granularity, such as the DAVIS-17 validation set, our model outperforms others by substantial margins of $1.2$, $1.2$, and $1.1$ on the \mjf, \mj, and \mf~scores, respectively. Such improvement highlights the effectiveness of structure refinement in capturing object details.
Moreover, OASIS remains stable performance gains on the MOSE benchmark, which contains more complex and densely packed objects that differ considerably from the training data. The contour accuracy (\mf) improves significantly by $1.0$, demonstrating that our approach enhances the structural integrity of segmentation, even in unseen and complex scenes.

Tab.~\ref{tab:mose-results} presents additional results where the training set includes the MOSE benchmark, allowing for a more comprehensive performance evaluation. In this setting, our approach still demonstrates superior results, significantly outperforming other methods on the DAVIS-2017 test set by margins of $1.7$, $1.7$, and $1.9$ on the \mjf, \mj, and \mf~scores, respectively. These results indicate that incorporating more complex scenes in the training set allows our model to develop a stronger structural understanding, leading to improved segmentation accuracy.

\begin{table}[!t]
    \centering
    \resizebox{\linewidth}{!}{
        \begin{tabular}{lccc||ccc}
        \toprule[1.5pt]  & \multicolumn{3}{c}{MOSE} & \multicolumn{3}{c}{DAV-17 Test} \\
        \cmidrule{2-7}
        Method & \mjf & \mj & \mf & \mjf & \mj & \mf \\
        \midrule[1pt]
        XMem \cite{xmem} & 59.6 & 55.4 & 63.7 & 79.6 & 76.1 & 83.0 \\
        DeAOT-R50 \cite{deaot} & 64.1 & 59.5 & 68.7 & 82.8 & 79.1 & 86.5 \\
        DEVA \cite{deva} & 66.0 & 61.8 & 70.3 & 82.6 & 78.9 & 86.4 \\
        Cutie \cite{cutie} & \boldblue{67.0} & \boldblue{62.8} & \boldblue{71.3} & \boldblue{83.7} & \boldblue{80.0} & \boldblue{87.4} \\
        \midrule
        \textbf{OASIS} & \textbf{67.5} & \textbf{63.2} & \textbf{71.9} & \textbf{85.4} & \textbf{81.7} & \textbf{89.3} \\
        \bottomrule[1.5pt]
        \end{tabular}}
    \caption{Performance comparison under Trained-with-MOSE setting on VOS benchmarks. Numbers are taken from \cite{cutie}. Boldface and bold blue text indicate the best and second-best performance in each column, respectively.
    }
    \label{tab:mose-results}
\end{table}

\subsubsection{Qualitative Experiments}
We illustrate the visual results of our method and the SOTA one on two typical video clips from the YouTubeVOS 2019 validation set in Fig.~\ref{fig:listin}. It can be observed that the existing SOTA method often struggles to distinguish between objects, particularly in cases of occlusion or high similarity between objects or between foreground objects and background. For instance, it fails to separate the fish hiding behind the other fish (the $1-3$ rows). In contrast, due to the proposed evidential learning, our OASIS is able to model the structure of the overlapped fish better and accomplishes precise segmentation in the first clip.  
Additionally, we observe that previous methods exhibit accumulating errors over time when segmenting the antennae of a snail (rows $4$-$6$), due to their reliance on memory mechanisms. In contrast, our proposed structure refinement module corrects these inaccuracies, significantly reducing this issue.

\begin{table}[!t]
    \centering
     \resizebox{\columnwidth}{!}{
        \begin{tabular}{lcccccc}
        \toprule[1.5pt]
        & \multicolumn{3}{c}{DAV-17 Val} & \multicolumn{3}{c}{DAV-17 Test-dev} \\ 
        \cmidrule{2-7} 
        Method  & \mjf & \mj & \mf & \mjf & \mj & \mf \\ 
        \midrule[1pt]
        Baseline  & 84.8 & 81.6 & 88.0 & 79.9 & 76.0 & 83.7 \\
        $+$CaE  &  84.9 & 81.8 & 88.1 & 80.0 & 76.0 & 84.0 \\
        $+$CaE$+$SD  & 85.1 & 82.0 & 88.1 & 80.3 & 76.3 & 84.5 \\
        $+$CaE$+$SD$^{*}$ & 85.4 & 82.2 & 88.6 & 80.8 & 76.8 & 84.9 \\
        \textbf{OASIS} & \textbf{85.6} & \textbf{82.4} & \textbf{88.7} & \textbf{81.0} & \textbf{76.9} & \textbf{85.1}\\
        \bottomrule[1.5pt]
        \end{tabular}
        }
    \caption{Ablation experiments about key components on DAVIS-2017 validation set and test-dev set.}
    \label{tab:abla}
\end{table}

\subsection{Ablation Study}
\subsubsection{Ablation on Key Components}
OASIS involves three novel designs: the integration of rough edges detected by the Canny filter, structure decoder for refinement, and evidential learning loss for uncertainty quantification. We conduct ablation experiments to assess the individual contribution of each component, with an additional experiment on the object features fusion mechanism in the structure decoder. To this end, we build five experimental settings in total: $+$CaE refers to the baseline model with pixel feature integrated with the rough Canny edges as per Eq. \ref{eq:roughfusion}, whereas it does not include the structure decoder and evidential learning. As shown in Fig. \ref{fig:sensory}, $+$CaE$+$SD means the structure decoder is further applied on top of $+$CaE for the refined structure map, while $+$CaE$+$SD$^{*}$ applied the object memory feature fusion mechanism. The final OASIS further adds uncertainty learning upon the $+$CaE$+$SD$^{*}$. Due to limited resources, all models are trained on the DAVIS-2017 training set with limited iterations and evaluated on the DAVIS-2017 validation and test-dev sets.

Tab.~\ref{tab:abla} shows the results of our ablation study. As observed on these benchmarks, Canny edge feature integration and the structure decoder are effective in characterizing the object and thus contribute to the overall segmentation performance. Object fusion further improves the capabilities of structure decoder, \emph{i.e.}, $+$CaE$+$SD$^{*}$ outperforms the one without object fusion largely especially on the test-dev set, with all indicators increased by about 0.5. Furthermore, the further introduction of uncertainty estimation and minimization allows our complete model to attain the best performance with negligible additional expense.

\begin{figure}[!t]
    \centering
    \includegraphics[width=0.9\linewidth]{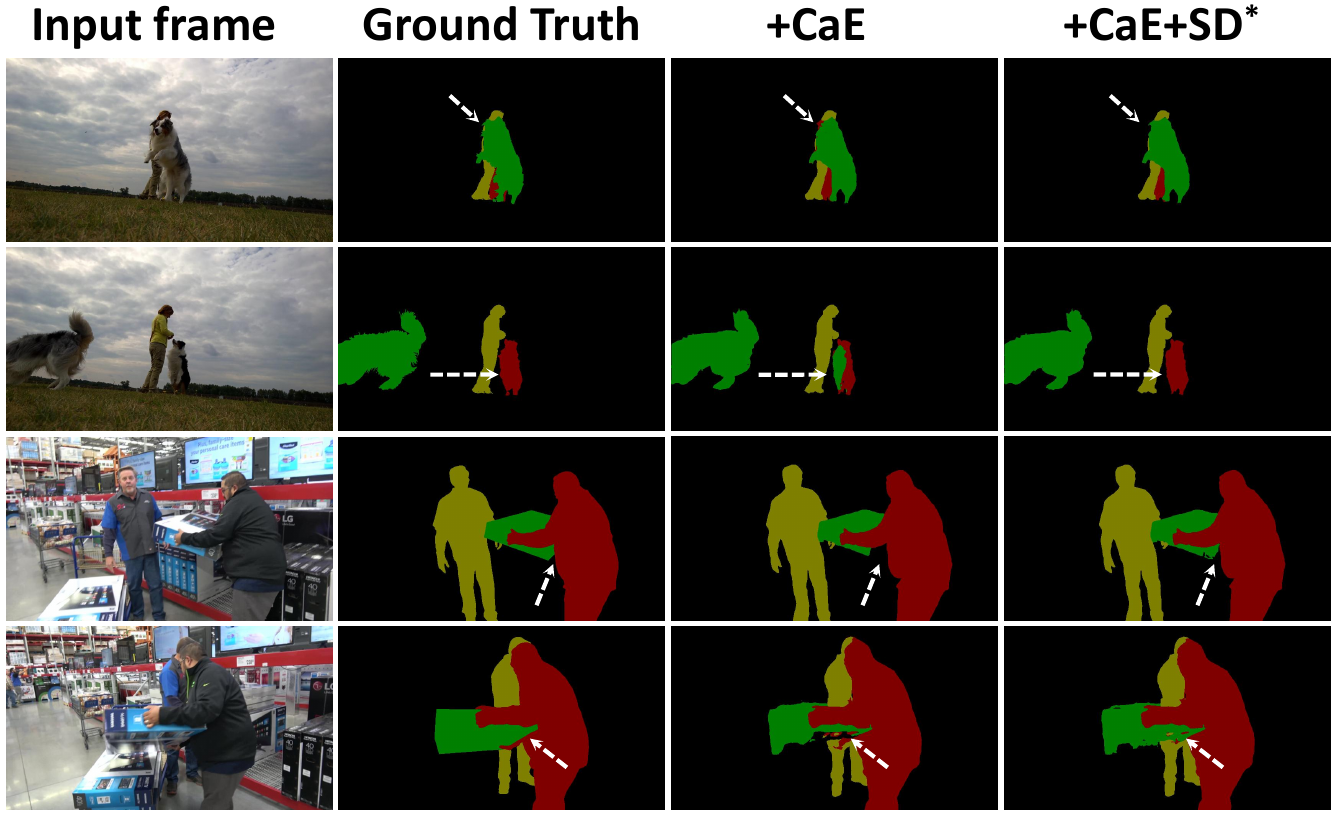}
    \caption{Qualitative comparison from different settings in ablating key components. We show the results of four typical frames. Zoom in for a better view.}
    \label{fig:abla-demo}
\end{figure}

\subsubsection{Visual Comparison on Key Components}
Fig.~\ref{fig:abla-demo} presents the visual results of different settings on representative frames. It can be observed that although $+$CaE achieves decent performance, structure refinement is still necessary to enhance output quality, especially for occlusions. For instance, at the position indicated by the white arrow (\emph{i.e.}, the body of the dog and the corner of the box), the implementation of the structure decoder significantly improves the structural details. This observation further validates the effectiveness of our design.

\subsubsection{Ablation on Factor $\epsilon$ and $\beta$}
In Eq.~\ref{eq:roughfusion} and Eq.~\ref{eq:refine}  we use factors $\epsilon$ and $\beta$ to control the impact brought by the rough edges and the refined structure map, respectively. We separately investigate the effect of different values for these two factors on model performance, with results presented in Tab.~\ref{tab:mapfactor}. Notably, as the two factors increase from $0$, accuracy consistently improves, substantiating the benefit of emphasizing boundary features. However, we observe that this performance gain reaches an upper limit, beyond which excessively large factors introduce a substantial distributional shift in the features, ultimately degrading performance. The best results are achieved when $\beta$ is greater than $\epsilon$, further underscoring the importance of more detailed structural information.

\begin{table}[!t]
    \centering
    \resizebox{1\columnwidth}{!}{
        \begin{tabular}{cccc||cccc}
            \toprule[1.5pt]
            & \multicolumn{3}{c||}{DAVIS-17 Val} & &\multicolumn{3}{c}{DAVIS-17 Val}\\ 
            \cmidrule{2-4} \cmidrule{6-8} 
            $\beta$ & \mjf & \mj & \mf &  $\epsilon$ & \mjf & \mj & \mf \\ 
            \midrule[0.25pt]
            $0.5$  & 85.3 & 82.1 & 88.5 & 0.0 & 85.3 & 82.2 & 88.5 \\ 
            $1.0$  & 85.6 & 82.4 & 88.7 & 0.5 & \textbf{85.6} & \textbf{82.4} & \textbf{88.7} \\ %
            $2.0$  & \textbf{85.7} & \textbf{82.5} & \textbf{88.9} & 1.0 & 85.1 & 81.8 & 88.3\\ 
            $5.0$  & 85.0 & 82.0 & 87.9 & 2.0 & 85.0 & 82.0 & 87.9\\   
            \bottomrule[1.5pt]
        \end{tabular}
    }
    \caption{Ablation experiments about factors $\epsilon$ and $\beta$ on the DAVIS-2017 validation set.}
    \label{tab:mapfactor}
\end{table}

\subsection{Performance Gain and Model Parameters}
In deep learning-based SVOS methods, performance gains may result from an increase in model parameters. In our approach, the primary source of additional parameters is the structure decoder, which introduces approximately 2M learnable parameters. Compared with the whole framework, these additional parameters are almost negligible.
To further validate that the effectiveness of OASIS does not primarily stem from the increased number of parameters, we conduct additional analytical experiments. To provide a fair comparison, we add a convolutional projection network of approximately 2M parameters to the baseline as a comparative method, while maintaining other settings and hyperparameters. Results are shown in Tab.~\ref{tab:para}. Notably, the method with a 2M-parameter network does not achieve better results than OASIS. Combined with the ablative results in Tab.~\ref{tab:abla}, which compare model performance with and without the structure decoder, we conclude that the proposed method is highly effective and, more importantly, that its performance gain is not due to the increased parameter count.
\begin{table}[!t]
    \centering
        \begin{tabular}{lccccc}
        \toprule[1.5pt]
        & &\multicolumn{4}{c}{DAVIS-17 Val} \\ 
        \cmidrule{3-6} 
        Method & \#Params & \mjf & \mj & \mf & FPS \\ 
        \midrule[0.25pt]
        Baseline  & 29M & 84.8 & 81.6 & 88.0 & \textbf{52} \\
        $+$2M Proj Net  & 31M & 85.1 & 82.0 & 88.1 & 37 \\ \midrule[0.5pt]
        OASIS  & 31M & \textbf{85.6} & \textbf{82.4} & \textbf{88.7} & 48 \\
        \bottomrule[1.5pt]
        \end{tabular}
    \caption{Experiments on the effect of including additional parameters. Results on the DAVIS-2017 validation set are shown.}
    \label{tab:para}
\end{table}

\section{Conclusion}
In this work, we propose a novel model named OASIS to address the challenge of occlusion in SVOS. OASIS first generates a structure map that captures the structural and layering information of predicted targets and leverages it to refine features, thus highlighting boundary regions. Additionally, we integrate evidential learning for uncertainty estimation to enhance the capability of OASIS to focus on complex and uncertain segmentation cases. Extensive experiments show the superior performance of OASIS while maintaining high computational efficiency. 

\appendix
\section*{Acknowledgments}
This work was supported by Ministry of Education Tier 2 grant, NUS, Singapore (T2EP20224-0028); and Ministry of Education Tier 1 grant, NUS, Singapore (24-1250-P0001).

{
    \small
    \bibliographystyle{IEEEtran}
    \bibliography{main}
}

\end{document}